\documentclass{svproc}
\usepackage{graphicx}%
\usepackage{multirow}%
\usepackage{amsmath,amssymb,amsfonts}%
\usepackage{mathrsfs}%
\usepackage[title]{appendix}%
\usepackage{xcolor}%
\usepackage{textcomp}%
\usepackage{manyfoot}%
\usepackage{booktabs}%
\usepackage{algorithm}%
\usepackage{algorithmicx}%
\usepackage{algpseudocode}%
\usepackage{listings}%
\usepackage{url}

\newlength\ColSpacing
\usepackage[usestackEOL]{stackengine}
\edef\tmp{\the\baselineskip}
\setstackgap{L}{\tmp}

\definecolor{blue}{rgb}{0, 0, 1}

\definecolor{red}{rgb}{1, 0, 0}

\makeatletter
\algrenewcommand\ALG@beginalgorithmic{\ttfamily}
\makeatother
\usepackage{etoolbox}\AtBeginEnvironment{algorithmic}{\scriptsize}

\usepackage{tabularx}

\begin{document}
\mainmatter              % start of a contribution
\title{Uncertainty in GNN Learning Evaluations: The Importance of a Consistent Benchmark for Community Detection.}
\titlerunning{Uncertainty in GNN Learning Evaluations}  % abbreviated title (for running head)
\author{William Leeney\inst{1} \and Ryan McConville\inst{2}}
\authorrunning{William Leeney et al.} % abbreviated author list (for running head)
\institute{University of Bristol\inst{1,2}\\
% \texttt{http://github.com/willleeney/ugle} \\
\email{will.leeney@bristol.ac.uk}}

\maketitle              % typeset the title of the contribution

\begin{abstract}

Graph Neural Networks (GNNs) have improved unsupervised community detection of clustered nodes due to their ability to encode the dual dimensionality of the connectivity and feature information spaces of graphs. Identifying the latent communities has many practical applications from social networks to genomics. Current benchmarks of real world performance are confusing due to the variety of decisions influencing the evaluation of GNNs at this task. To address this, we propose a framework to establish a common evaluation protocol. We motivate and justify it by demonstrating the differences with and without the protocol. The W Randomness Coefficient is a metric proposed for assessing the consistency of algorithm rankings to quantify the reliability of results under the presence of randomness. We find that by ensuring the same evaluation criteria is followed, there may be significant differences from the reported performance of methods at this task, but a more complete evaluation and comparison of methods is possible.


\end{abstract}

%%%%%%%%%%%%%%%%%%%%%%%%%%%%%%%%%%%%%%%%%%%%%%%%%%%%%%%%%%%%%%%%%%%%%%%%%%%%%%%%%%%%%%%%
%%%%%%%%%%%%%%%%%%%%%%%%%%%%%%%%%%%% INTRODUCTION %%%%%%%%%%%%%%%%%%%%%%%%%%%%%%%%%%%%%%
%%%%%%%%%%%%%%%%%%%%%%%%%%%%%%%%%%%%%%%%%%%%%%%%%%%%%%%%%%%%%%%%%%%%%%%%%%%%%%%%%%%%%%%%

\section{Introduction}
\vspace{-1mm}

%%%%% graph neural networks and why they are useful 
GNNs are a popular neural network based approach for processing graph-structured data due to their ability to combine two sources of information by propagating and aggregating node feature encodings along the network's connectivity \cite{kipf2016semi}. Nodes in a network can be grouped into communities based on similarity in associated features and/or edge density \cite{schaeffer2007graph}. Analysing the structure to find clusters, or communities, of nodes provides useful information for real world problems such as  misinformation detection \cite{monti2019fake}, genomic feature discovery \cite{cabreros2016detecting}, social network or research recommendation \cite{yang2013community}. As an unsupervised task, clusters of nodes are identified based on the latent patterns within the dataset, rather than ``ground-truth" labels. Assessing performance at the discovery of unknown information is useful to applications where label access is prohibited. Some applications of graphs deal with millions of nodes and there is a low labelling rate with datasets that mimic realistic scenarios \cite{hu2021ogb}. In addition, clustering is particularity relevant for new applications where there is not yet associated ground truth. 

However, there is no widely accepted or followed way of evaluating algorithms that is done consistently across the field, despite benchmarks being widely considered as important. Biased benchmarks, or evaluation procedures, can mislead the narrative of research which distorts understanding of the research field. Inconclusive results that may be valuable for understanding or building upon go unpublished, which wastes resources, money, time and energy that is spent on training models. In fields where research findings inform policy decisions or medical practices, publication bias can lead to decisions based on incomplete or biased evidence, potentially causing harm or inefficiency. To accurately reflect the real-world capabilities of research, it would beneficial to use a common framework for evaluating proposed methods.

%%%%%%%%%%%%%%%% 
The framework detailed herein is a motivation and justification for this position. To demonstrate the need for this, we measure the difference between using the default parameters given by the original implementations to those optimised for under this framework. A metric is proposed for evaluating consistency of algorithm rankings over different random seeds which quantifies the robustness of results. This work will help guide practitioners to better model selection and evaluation within the field of GNN community detection.

\textbf{Contributions}
In this paper, we make the following key contributions: 1) We demonstrate that despite benchmarks being accepted as important, many experiments in the field follow different, often unclear, benchmarking procedures to report performance  2) We propose a framework for improving and producing fairer comparisons of GNNs at the task of clustering. For enablement, we open source the code\footnote{\url{https://github.com/willleeney/ugle}} and encourage model developers to submit their methods for inclusion. 3) We quantify the extent to which a hyperparameter optimisation and ranking procedure affects performance using the proposed W Randomness Coefficient, demonstrating that model selection requires use of the framework for accurate comparisons.
 
%%%%%%%%%%%%%%%%%%%%%%%%%%%%%%%%%%%%%%%%%%%%%%%%%%%%%%%%%%%%%%%%%%%%%%%%%%%%%%%%%%%%%%%%
%%%%%%%%%%%%%%%%%%%%%%%%%%%%%%%%%%% Related Work %%%%%%%%%%%%%%%%%%%%%%%%%%%%%%%%%%%%%%%
%%%%%%%%%%%%%%%%%%%%%%%%%%%%%%%%%%%%%%%%%%%%%%%%%%%%%%%%%%%%%%%%%%%%%%%%%%%%%%%%%%%%%%%%

\section{Related Work}
\vspace{-1mm}

There is awareness of need for rigour in frameworks evaluating machine learning algorithms \cite{pineau2021improving}.
Several frameworks for the evaluation of supervised GNNs performing node classification, link prediction and graph classification exist \cite{dwivedi2020benchmarking,morris2020tudataset,errica2019fair,palowitch2022graphworld}. 
We are interested in unsupervised community detection which is harder to train and evaluate. 
Current reviews of community detection provide an overview but do not evaluate and the lack of a consistent evaluation at this task has been discussed for non-neural methods \cite{liu2020deep,jin2021survey} but this doesn't currently include GNNs. 

There are various frameworks for assessing performance and the procedure used for evaluation changes the performance of all algorithms \cite{zoller2021benchmark}. Under consistent conditions, it has been show that simple models can perform better with a thorough exploration of the hyperparameter space \cite{shchur2018pitfalls}. This can be because performance is subject to random initialisations \cite{errica2019fair}. Lifting results from papers without carrying out the same hyperparameter optimisation over all models is not consistent and is a misleading benchmark. Biased selection of random seeds that skew performance is not fair. Not training over the same number of epochs or not implementing model selection based on the validation set results in unfair comparisons with inaccurate conclusions about model effectiveness. Hence, there is no sufficient empirical evaluation of GNN methods for community detection as presented in this work.

%%%%%%%%%%%%%%%%%%%%%%%%%%%%%%%%%%%%%%%%%%%%%%%%%%%%%%%%%%%%%%%%%%%%%%%%%%%%%%%%%%%%%%%%
%%%%%%%%%%%%%%%%%%%%%%%%%%%%%%%%%%% METHODOLOGY %%%%%%%%%%%%%%%%%%%%%%%%%%%%%%%%%%%%%%%%
%%%%%%%%%%%%%%%%%%%%%%%%%%%%%%%%%%%%%%%%%%%%%%%%%%%%%%%%%%%%%%%%%%%%%%%%%%%%%%%%%%%%%%%%

\section{Methodology}
\vspace{-1mm}

This section details the procedure for evaluation; the problem that is aimed to solve; the hyperparameter optimisation and the resources allocated to this investigation; the algorithms that are being tested; the metrics of performance and datasets used. At the highest level, the framework coefficient calculation is summarised by \textbf{Algorithm 1}. 

\setlength{\intextsep}{10pt}% Remove \textfloatsep
\begin{algorithm}[h]
    \caption{Overview of the evaluation framework. \textit{train()} is the function to train a model; \textit{optimise()} encompasses the hyperparameter optimisation and uses the model, training function, resources allocated and the current evaluation test;  \textit{evaluate()} returns the performance of a model and \textit{ranking-coefficient()} calculates the rankings and coefficient of agreement across the randomness over all tests.  \label{algo: framework}} 
	\begin{algorithmic}
        \Require tests, models, resources, \textit{train(), optimise() evaluate(), ranking-coefficient()}
        \For {test in tests}:
            \For {model in models}:
                \State \textit{optimise}(model, resources, test, \textit{train()})
                \State results += \textit{evaluate}(model, resources, test)
            \EndFor
        \EndFor 
        \State W = \textit{ranking-coefficient}(results)
	\end{algorithmic}
\end{algorithm}

The current approach to evaluation of algorithms suffer from several shortcomings that hinder the fairness and reliability of model comparisons. A consistent framework establishes a clear and objective basis for comparing models. A common benchmark practice promotes transparency by comprehensively documenting what affects performance, encouraging researchers to compare fairly. Only if the exact same evaluation procedure is followed can results be lifted from previous research, which saves the time and energy of all practitioners. A consistent practise must be established as there is current no reason for confidence in claims of performance. Trustworthy results builds understanding and allows progression of the field.

To evaluate the consistency and quality of the results obtained by this framework, two metrics are proposed. No algorithm evaluated is deterministic as all require randomness to initialise the network. Therefore, the consistency of a framework can be considered as the average amount that the performance rankings change when different randomness is present over all tests in the framework. The tests of a framework are the performance of a metric on a specific dataset, the ranking is where an algorithm places relative to the other algorithms. Here, the different randomness means that each random seed that the algorithms are evaluated on. 

Kendall's $W$ coefficient of concordance \cite{field2005kendall} is used as the basis for the consistency metric. This is a statistical measure used to assess the level of agreement among multiple judges or raters who rank or rate a set of items. It quantifies the degree of agreement among the judges' rankings. The coefficient ranges from 0 to 1, where higher values indicate greater concordance or agreement among the rankings. The formula for calculating Kendall's $W$ involves comparing pairs of items for each judge and computing the proportion of concordant pairs. We adapt the coefficient to quantify how consistent the ranking of the algorithms under this framework and refer to it as the $W$ Randomness Coefficient. This can be calculated with a number of algorithms $a$, and random seeds $n$, along with tests of performance that create rankings of algorithms. For each test of performance, Kendall's $W$ coefficient of concordance is calculated for the consistency of rankings across the random seed. The sum of squared deviations $S$ is calculated for all algorithms, and calculated using the deviation from mean rank due for each random seed. This is averaged over all metrics and datasets that make up the suite of tests $\mathcal{T}$ to give the $W$ Randomness Coefficient defined as Equation \ref{equ: Worder}. Using the one minus means that if the $W$ is high then randomness has affected the rankings, whereas a consistent ranking procedure results in a lower number. By detailing the consistency of a framework across the randomness evaluated, the robustness of the framework can be maintained, allowing researchers to trust results and maintain credibility of their publications. 

\begin{equation}
    W = 1 - \frac{1}{|\mathcal{T}|}\sum_{t \in \mathcal{T}} \frac{12S}{n^2(a^3-a)}
    \label{equ: Worder}
\end{equation}

The second metric is used to compare whether the performance is better under the hyperparameters or default reported by original implementations. The different parameter sets are given a rank by comparing the performance on every test. This is then averaged across every test, to give the framework comparison rank. Demonstrating that failing to optimize hyperparameters properly can result in sub-optimal performance means that models that could have performed better with proper tuning may appear inferior. This affects decision-making and potentially leading to the adoption of sub-optimal solutions. In the real world, this can have costly and damaging consequences, and is especially critical in domains where model predictions impact decisions, such as healthcare, finance, and autonomous vehicles.

\subsection{Problem Definition}\label{section: problem_def}

The problem definition of community detection on attributed graphs is defined as follows. The graph, where $N$ is the number of nodes in the graph, is represented as $G = (A, X)$, with the relational information of nodes modelled by the adjacency matrix $A \in \mathbb{R}^{N \times N}$. Given a set of nodes $V$ and a set of edges $E$, let $e_{i, j} = (v_i, v_j) \in E$ denote the edge that points from $v_j$ to $v_i$. The graph is considered weighted so, the adjacency matrix $ 0 < A_{i, j} \leq 1$ if $e_{i,j} \in E$  and $A_{i, j} = 0$ if $e_{i,j} \notin E$. Also given is a set of node features $X \in \mathbb{R}^{N \times d} $, where $d$ represents the number of different node attributes (or feature dimensions). The objective is to partition the graph $G$ into $k$ clusters such that nodes in each partition, or cluster, generally have similar structure and feature values. The only information typically given to the algorithms at training time is the number of clusters $k$ to partition the graph into. Hard clustering is assumed, where each community detection algorithm must assign each node a single community to which it belongs, such that $P \in \mathbb{R}^{N}$ and we evaluate the clusters associated with each node using the labels given with each dataset such that $L \in \mathbb{R}^{N}$.

\subsection{Hyperparameter Optimisation Procedure}

There are sweet spots of architecture combinations that are best for each dataset \cite{bergstra2012random} and the effects of not selecting hyperparameters (HPs) have been well documented. Choosing too wide of a HP interval or including uninformative HPs in the search space can have an adverse effect on tuning outcomes in the given budget \cite{yang2020hyperparameter}. Thus, a HPO is performed under feasible constraints in order to validate the hypothesis that HPO affects the comparison of methods. It has been shown that grid search is not suited for searching for HPs on a new dataset and that Bayesian approaches perform better than random \cite{bergstra2012random}. There are a variety of Bayesian methods that can be used for hyperparameter selection. One such is the Tree Parzen-Estimator (TPE) \cite{bergstra2013making} that can retain the conditionality of variables \cite{yang2020hyperparameter} and has been shown to be a good estimator given limited resources \cite{yuan2021systematic}. The multi-objective version of the TPE \cite{ozaki2020multiobjective} is used to explore the multiple metrics of performance investigated. Given a limited budget, the TPE is optimal as it allows the efficient exploration of parameters.

\setlength\tabcolsep{7pt}
\begin{table}[htpb]
\centering
\caption{Resources are allocated an investigation, those detailed are shared across all investigations. Algorithms that are designed to benefit from a small number of HPs should perform better as they can search more of the space within the given budget. All models are trained with 1x 2080 Ti GPU on a server with 12GB of RAM, and a 16core Xeon CPU. }\label{tab: resources} %
\begin{tabular}{@{}c|c@{}}
\toprule
resource & associated allocation  \\
\midrule
\midrule
optimiser & Adam \\
learning rate & $\{0.05, 0.01, 0.005, 0.001, 0.0005, 0.0001 \}$ \\
weight decay & $\{0.05, 0.005, 0.0005, 0.0\}$ \\
max epochs & $5000$ \\
patience & $\{25, 100, 500, 1000\}$ \\
max hyperparameter trials & $300$ \\
seeds & $\{42, 24, 976, 12345, 98765, 7, 856, 90, 672, 785\}$ \\
training/validation split & $0.8$ \\ 
train/testing split & $0.8$ \\

\bottomrule
\end{tabular}
\end{table}
\noindent

In this framework, a modification to the nested cross validation procedure is used to match reasonable computational budgets, which is to optimise hyperparameters on the first seed tested on and use those hyperparameters on the other seeds. Additionally, it it beneficial to establish a common resource allocation such as number of epochs allowed for in training or number of hyperparameter trials. Ensuring the same resources are used in all investigations means that the relatively underfunded researchers can participate in the field, democratising the access to contribution. Conversely, this also means that highly funded researchers cannot bias their results by exploiting the resources they have available.

\subsection{Suite of Tests}

A test of an algorithm in this framework is the performance under a metric on a dataset. Algorithms are ranked on each test on every random seed used. For evaluation purposes, some metrics require the ground truth, and others do not, although regardless, this knowledge is not used during the training itself. Macro F1-score (F1) is calculated to ensure that evaluations are not biased by the label distribution, as communities sizes are often imbalanced. Normalised mutual information (NMI) is also used, which is the amount of information that can extract from one distribution with respect to a second. 

For unsupervised metrics, modularity and conductance are selected. Modularity quantifies the deviation of the clustering from what would be observed in expectation under a random graph. Conductance is the proportion of total edge volume that points outside the cluster. These two metrics are unsupervised as they are calculated using the predicted cluster label and the adjacency matrix, without using any ground truth. Many metrics are used in the framework as they align with specific objectives and ensures that evaluations reflect a clear and understandable assessment of performance. 

Generalisation of performance on one dataset can often not be statistically valid and lead to overfitting on a particular benchmark \cite{salzberg1997comparing}, hence multiple are used in this investigation. To fairly compare different GNN architectures, a range of graph topologies are used to fairly represent potential applications. Each dataset can be summarised by commonly used graph statistics: the average clustering coefficient \cite{watts1998collective} and closeness centrality \cite{wasserman1994social}. The former is the proportion of all the connections that exist in a nodes neighbourhood compared to a fully connected neighbourhood, averaged across all nodes. The latter is the reciprocal of the mean shortest path distance from all other nodes in the graph. All datasets are publicly available\footnote{\url{https://github.com/yueliu1999/Awesome-Deep-Graph-Clustering}}, and have been used previously in GNN research \cite{deep_graph_clustering_survey}. 

Using many datasets for evaluation means that dataset bias is mitigated, which means that the framework is better at assessing the generalisation capability of models to different datasets. These datasets are detailed in Table \ref{tab:stats}, the following is a brief summary: Cora \cite{mccallum2000automating}, CiteSeer \cite{giles1998citeseer}, DBLP \cite{tang2008arnetminer} are graphs of academic publications from various sources with the features coming from words in publications and connectivity from citations. AMAC and AMAP are extracted from the Amazon co-purchase graph \cite{he2016ups}. Texas, Wisc and Cornell are extracted from web pages from computer science departments of various universities \cite{craven1998learning}. UAT, EAT and BAT contain airport activity data collected from the National Civil Aviation Agency, Statistical Office of the European Union and Bureau of Transportation Statistics \cite{deep_graph_clustering_survey}.

\setlength\tabcolsep{3.5pt}
\begin{table}[h]
\centering
\caption{The datasets and associated statistics.}\label{tab:stats}%
\begin{tabular}{@{}c|cccccc@{}}
\toprule
\textbf{\Longunderstack{\\Datasets}}  & \Longunderstack{\\Nodes} & \Longunderstack{\\Edges} & \Longunderstack{\\Features} & \Longunderstack{\\Classes} & \Longunderstack{Average\\Clustering\\ Coefficient} & \Longunderstack{Mean\\Closeness\\ Centrality} \\
\midrule
\midrule
amac \cite{he2016ups} & 13752 & 160124 & 767 & 10 & 0.157 & 0.264 \\
amap \cite{he2016ups} & 7650 & 238163 & 745 & 8 & 0.404 & 0.242 \\
bat \cite{deep_graph_clustering_survey} & 131 & 2077 & 81 & 4 & 0.636 & 0.469 \\
citeseer  \cite{giles1998citeseer} & 3327 & 9104 & 3703 & 6 & 0.141 & 0.045 \\
cora \cite{mccallum2000automating} & 2708 & 10556 & 1433 & 7 & 0.241 & 0.137 \\
dblp \cite{tang2008arnetminer} & 4057 & 7056 & 334 & 4 & 0.177 & 0.026 \\
eat \cite{deep_graph_clustering_survey} & 399 & 11988 & 203 & 4 & 0.539 & 0.441 \\
uat \cite{deep_graph_clustering_survey} & 1190 & 27198 & 239 & 4 & 0.501 & 0.332 \\
texas \cite{craven1998learning} & 183 & 325 & 1703 & 5 & 0.198 & 0.344 \\
wisc \cite{craven1998learning} & 251 & 515 & 1703 & 5 & 0.208 & 0.32 \\
cornell \cite{craven1998learning} & 183 & 298 & 1703 & 5 & 0.167 & 0.326 \\
\bottomrule
\end{tabular}
\end{table}
\noindent

\subsection{Models}

We consider a representative suite of GNNs, selected based on factors such as code availability and re-implementation time.
In addition to explicit community detection algorithms, we also consider those that can learn an unsupervised representation of data as there is previous research that applies vector-based clustering algorithms to the representations \cite{mcconville2021n2d}.

Deep Attentional Embedded Graph Clustering (DAEGC) uses a k-means target to self-supervise the clustering module to iteratively refine the clustering of node embeddings \cite{wang2019attributed}. Deep Modularity Networks (DMoN) uses GCNs to maximises a modularity based clustering objective to optimise cluster assignments by a spectral relaxation of the problem \cite{tsitsulin2020graph}. Neighborhood Contrast Framework for Attributed Graph Clustering (CAGC) \cite{NCAGC} is designed for attributed graph clustering with contrastive self-expression loss that selects positive/negative pairs from the data and contrasts representations with its k-nearest neighbours. Deep Graph Infomax (DGI) maximises mutual information between patch representations of sub-graphs and the corresponding high-level summaries \cite{velickovic2019deep}. GRAph Contrastive rEpresentation learning (GRACE) generates a corrupted view of the graph by removing edges and learns node representations by maximising agreement across two views \cite{zhu2020deep}. Contrastive Multi-View Representation Learning on Graphs (MVGRL) argues that the best employment of contrastive methods for graphs is achieved by contrasting encodings' from first-order neighbours and a general graph diffusion \cite{hassani2020contrastive}. Bootstrapped Graph Latents (BGRL) \cite{thakoor2021bootstrapgraph} uses a self-supervised bootstrap procedure by maintaining two graph encoders; the online one learns to predict the representations of the target encoder, which in itself is updated by an exponential moving average of the online encoder. SelfGNN \cite{kefato2021selfgnn} also uses this principal but uses augmentations of the feature space to train the network. Towards Unsupervised Deep Graph Structure Learning (SUBLIME) \cite{liu2022towards} an encoder with the bootstrapping principle applied over the feature space as well as a contrastive scheme between the nearest neighbours. Variational Graph AutoEncoder Reconstruction (VGAER) \cite{qiu2022VGAER} reconstructs a modularity distribution using a cross entropy based decoder from the encoding of a VGAE \cite{kipf2016variational}.

%%%%%%%%%%%%%%%%%%%%%%%%%%%%%%%%%%%%%%%%%%%%%%%%%%%%%%%%%%%%%%%%%%%%%%%%%%%%%%%%%%%%%%%%
%%%%%%%%%%%%%%%%%%%%%%%%%%%%%%%%%%%%%% RESULTS %%%%%%%%%%%%%%%%%%%%%%%%%%%%%%%%%%%%%%%%%
%%%%%%%%%%%%%%%%%%%%%%%%%%%%%%%%%%%%%%%%%%%%%%%%%%%%%%%%%%%%%%%%%%%%%%%%%%%%%%%%%%%%%%%%
\section{Evaluation and Discussion}
\vspace{-1mm}

The Framework Comparison Rank is the average rank when comparing performance of the parameters found through hyperparameter optimisation versus the default values. From Table \ref{tab: metric_stats} it can be seen that Framework Comparison Rank indicates that the hyperparameters that are optimised on average perform better. This validates the hypothesis that the hyperparameter optimisation significantly impacts the evaluation of GNN based approaches to community detection. The $W$ Randomness Coefficient quantifies the consistency of rankings over the different random seeds tested on, averaged over the suite of tests in the framework. With less deviation of prediction under the presence of randomness, an evaluation finds a more confident assessment of the best algorithm. A higher $W$ value using the optimised hyperparameters indicates that the default parameters are marginally more consistent over randomness however this does deviate more across all tests. By using and optimising for the $W$ Randomness Coefficient with future extensions to this framework, we can reduce the impact of biased evaluation procedures. With this coefficient, researchers can quantify how trustworthy their results are, and therefore the usability in real-world applications. It is likely that there is little difference in the coefficients in this scenario as the default parameters have been evaluated with a consistent approach to model selection and constant resource allocation to training time. This sets the baseline for consistency in evaluation procedure and allows better understanding of relative method performance.

\setlength\tabcolsep{7pt}
\begin{table}[htbp]
\centering
\caption{Here is the quantification of intra framework consistency using the $W$ Randomness Coefficient and inter framework disparity using the Framework Comparison Rank. Low values for the Framework Comparison Rank and $W$ Randomness Coefficient are preferred. }\label{tab: metric_stats}%
\begin{tabular}{@{}c|cc@{}}
\toprule
Results & Default & HPO \\
\midrule
\midrule
\Longunderstack{$W$ Randomness\\ Coefficient} & $0.476\pm 0.200$ & $0.489\pm0.144$ \\
\midrule
\midrule
\Longunderstack{Framework \\ Comparison Rank} & $1.829\pm0.24$ & $1.171\pm0.24$ \\

\end{tabular}
\end{table}

\begin{figure}[H]
    \centering
\includegraphics[width=\textwidth,height=0.9\textheight,keepaspectratio]{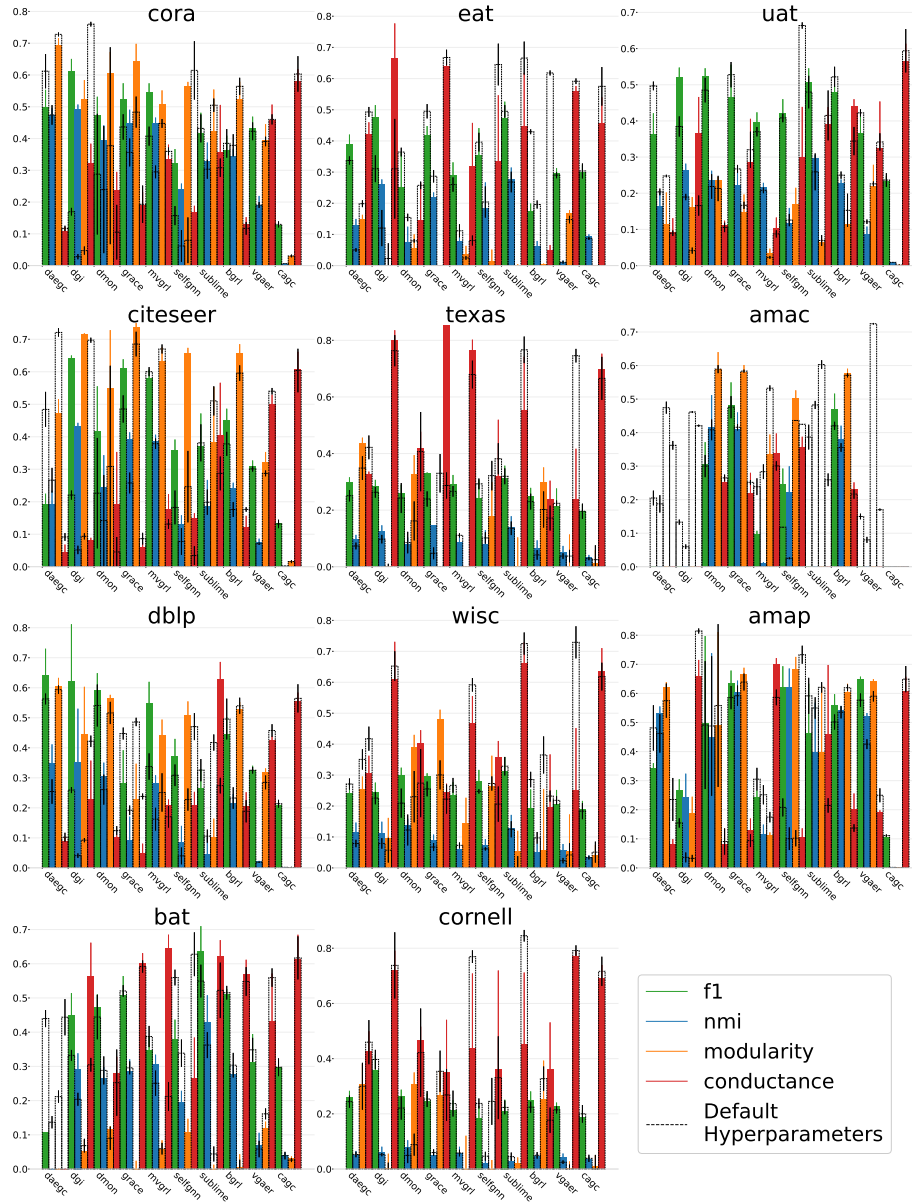}
    \caption{The average performance and standard deviation of each metric averaged over every seed tested on for all methods on all datasets. The hyperparameter investigation under our framework is shown in colour compared with the default hyperparameters in dashed boxes. The lower the value for Conductance is better. Out Of Memory (OOM) occurrences on happened on the amac dataset with the following algorithms during HPO: daegc, sublime, dgi, cagc, vgaer and cagc under the default HPs. } 
    \label{fig: results_fig}
\end{figure}

The results of the hyperparameter optimisation and the default parameters is visualised in Figure \ref{fig: results_fig}. From this, we can see the difference in performance under both sets of hyperparameters. On some datasets the default parameters work better than those optimised, which means that under the reasonable budget assumed in this framework they are not reproducible. This adds strength to the claim that using the default parameters is not credible. Without validation that these parameters can be recovered, results cannot be trusted and mistakes are harder to identify. 
On the other side of this, often the hyperparameter optimisation performs better than the default values. Algorithms were published knowing these parameters can be tuned to better performance. Using a reasonable resources, the performance can be increased, which means that without the optimisation procedure, inaccurate or misleading comparisons are propagated. Reproducible findings are the solid foundation that allows us to build on previous work and is a necessity for scientific validity. 

Given different computational resources performance rankings will vary. Future iterations of the framework should experiment with the the number of trials and impact of over-reliance on specific seeds or extending the hyperparameter options to a continuous distribution. Additionally, finding the best general algorithm will have to include a wide range of different topologies or sizes of graphs that are not looked at, neither do we explore other feature space landscapes or class distributions.

\section{Conclusion}
\vspace{-1mm}

In this work we demonstrate flaws with how GNN based community detection methods are currently evaluated, leading to potentially misleading and confusing conclusions. To address this, an evaluation framework is detailed for evaluating GNNs at community detection that provides a more consistent and fair evaluation, and can be easily extended. We provide further insight that consistent HPO is key at this task by quantifying the difference in performance from HPO to reported values. Finally, a metric is proposed for the assessing the consistency of rankings that empirically states the trust researchers can have in the robustness of results.

\bibliographystyle{spmpsci} % We choose the "plain" reference style
\bibliography{refs} % Entries are in the refs.bib file
\end{document}